\documentclass[sn-mathphys,Numbered]{sn-jnl}
\usepackage{graphicx}%
\usepackage{multirow}%
\usepackage{amsmath,amssymb,amsfonts}%
\usepackage{amsthm}
\usepackage{mathrsfs}%
\usepackage[title]{appendix}%
\usepackage{xcolor}%
\usepackage{textcomp}%
\usepackage{manyfoot}%
\usepackage{booktabs}%
\usepackage{algorithm}%
\usepackage{algorithmicx}%
\usepackage{algpseudocode}%
\usepackage{listings}%
\usepackage{titlesec}
\usepackage{pifont}
\theoremstyle{thmstyleone}%
\theoremstyle{thmstyletwo}%

\theoremstyle{thmstylethree}%

\begin{document}

\title[Article Title]{Editing Implicit and Explicit Representations of Radiance Fields: A Survey}


\author*[1]{\fnm{Arthur} \sur{Hubert}}\email{arthur.hubert@uni.lu}

\author[1]{\fnm{Gamal} \sur{Elghazaly}}\email{gamal.elghazaly@uni.lu}

\author[1]{\fnm{Raphael} \sur{Frank}}\email{raphael.frank@uni.lu}

\affil[1]{\orgname{SnT - Interdisciplinary Centre for Security, Reliability and Trust, University of Luxembourg}, \orgaddress{\street{29 Avenue John F. Kennedy}, \city{Luxembourg}, \postcode{1855}, \country{Luxembourg}}}


\abstract{Neural Radiance Fields (NeRF) revolutionized novel view synthesis in recent years by offering a new volumetric representation, which is compact and provides high-quality image rendering. However, the methods to edit those radiance fields developed slower than the many improvements to other aspects of NeRF. With the recent development of alternative radiance field-based representations inspired by NeRF as well as the worldwide rise in popularity of text-to-image models, many new opportunities and strategies have emerged to provide radiance field editing. In this paper, we deliver a comprehensive survey of the different editing methods present in the literature for NeRF and other similar radiance field representations. We propose a new taxonomy for classifying existing works based on their editing methodologies, review pioneering models, reflect on current and potential new applications of radiance field editing, and compare state-of-the-art approaches in terms of editing options and performance.}




\maketitle

\section{Introduction}

With applications in augmented reality, medical imaging, photorealistic simulation, and more, high-quality 3D models have become increasingly valuable. However, 3D assets traditionally need to be created manually by highly skilled experts and artists. Many applications also require rigging to allow precise manipulation and deformation of such assets. In order to increase the quality of photorealistic 3D environments and make their creation easier, many have turned to automated methods to generate 3D representations of real-world objects \cite{tancik2022block, fei20243d}.

The introduction of NeRF \cite{mildenhall2021nerf} in 2020 marked a breakthrough in 3D scene representation, offering a high-quality novel view synthesis method capable of reconstructing realistic scenes from a collection of posed images. Its neural representation provides a compact model with good results. Shortly thereafter, further work on the NeRF baseline greatly improved many of its weaknesses in terms of training performance \cite{muller2022instant}, visual quality \cite{barron2021mip}, scalability \cite{tancik2022block}, and handling of dynamic scenes \cite{pumarola2021d}. This field has gained a lot of traction in the four years since its appearance, with dozens of papers published every day. The evolution of NeRF led to many applications in medical imaging \cite{wang2024neural} or autonomous driving \cite{he2024neural}. However, an aspect that received less attention initially gained more traction recently: \textit{Radiance Field Editing}. Unlike other popular 3D representations~\cite{laine2010efficient}, NeRF models rely on an implicit neural network, which raises multiple challenges to edit them. Considering the many existing applications for NeRF and 3D assets, precise control of objects and scenes represented by these methods provides a unique opportunity. Basic NeRF representations are bound to the source images used to train them, making it difficult to create new models or control represented scenes. 

Following the popularity of NeRF, other photorealistic radiance field-based methods emerged to compensate for its shortcomings. One promising approach is 3D Gaussian Splatting (3DGS)~\cite{3DGS}, which replaces the neural network component of the NeRF with a large collection of 3D Gaussian Ellipsoids with individual color values. This representation offers several advantages over the NeRF approach: (1) it enables significantly faster training and rendering, (2) provides a directly explicit structure, and (3) facilitates various tasks, including the editing. 

In this paper, we propose a comprehensive survey of the different strategies existing in the literature to edit radiance field representations, with a strong focus on NeRF-based and 3DGS-based strategies, the two foundational approaches underlying most current research. In Section \ref{sec:related}, we first cover related surveys and how they differ from the present review. We then introduce the basic concepts of radiance fields in Section \ref{sec:background}, before proposing a taxonomy of these different editing methods in Section \ref{sec:taxonomy}. In the same section, we also provide a detailed explanation of how these methods enable editing, highlighting pioneering models in the field, and reviewing the current state-of-the-art techniques. Although a large portion of this description focuses on the NeRF representations that introduced them, it is important to note that many of these methods are also adaptable to 3DGS. In Section \ref{sec:application} we discuss several application areas for radiance field editing followed by the introduction of popular datasets and evaluation metrics in Section \ref{sec:evaluation}. In Section \ref{sec:future_works}, we present and discuss future research challenges, before we conclude the survey in Section \ref{sec:conclusion}.

\section{Related Work} \label{sec:related}

Several survey papers on NeRFs \cite{gao2022nerf, zhu2023deep, rabby2023beyondpixels, cai2024nerf} and 3DGS \cite{luo2024review, chen2024survey, wu2024recent, fei20243d, bao20243d, dalal2024gaussian, bagdasarian20243dgs} have recently underscored the major advances in real-time rendering, 3D representation \cite{gao2022nerf}, and interactive applications in 3D computer graphics \cite{gao2022nerf}, computer vision \cite{gao2022nerf}, robotics \cite{tosi2024nerfs, wang2024nerf, yang2024slam}, virtual reality and autonomous navigation \cite{he2024neural, ming2024benchmarking}. These works collectively illustrate how NeRFs and 3DGS enable unprecedented fidelity in rendering and adaptability in various applications, bringing efficiency and real-time capabilities \cite{yan2024neural} to complex, dynamic, and interactive environments.

\subsection{Surveys on NeRFs}

One of the first survey papers on NeRFs is the work in \cite{gao2022nerf} which provides a detailed review of recent advancements in NeRF models, with a taxonomy more focused on their architecture and applications, while providing a more thorough introduction to the theory of NeRFs and the underlying training through differentiable volume rendering. However, the taxonomy provided is limited to a selected number of key NeRF papers. Alternatively, a deeper review and analysis of recent advances in NeRFs have been discussed in \cite{zhu2023deep}. The deep analysis behind this paper lies in a novel taxonomy based on the various characters of NeRFs and the associated breakthroughs among which factorizable embedded space and breakthroughs in relightable \cite{srinivasan2021nerv, boss2021nerd}, deformable \cite{park2021hypernerf} and scene editing \cite{graf, Editing_Conditional_RF, giraffe} in NeRFs. Unlike other survey papers, which mainly focused on conventional 3D computer vision approaches and early contributions in NeRFs, the comprehensive review in \cite{rabby2023beyondpixels} has primarily explored the full potential of NeRFs and discusses their limitations and capabilities. In the context of scene editing, \cite{rabby2023beyondpixels} has also discussed some scene editing approaches while discussing and comparing their potential and limitations. Other survey papers have focused on summarizing key algorithms and important works rather than applications of NeRFs \cite{cai2024nerf}.

%% Applications (robot navigation tasks a bit of 3DGS)
NeRFs have been explored to improve autonomous navigation in robotics, focusing on perception, localization, and decision-making in the survey \cite{ming2024benchmarking}.
The work also benchmarks NeRF-based methods in 3D reconstruction, SLAM, and navigation, discussing strengths, limitations, and future research avenues. Advanced techniques like 3DGS and generative AI integration are highlighted for their potential to improve scene understanding and efficiency.

Early works on NeRFs have focused on the reconstruction of static scenes. Although challenging, extending NeRFs to dynamic scenes has always been of interest for their potential in real-world applications. \cite{lin2024dynamic} has detailed and analyzed the evolution of dynamic NeRFs as well as key methods and principles to implement dynamic NeRFs. As dynamic NeRFs share similar features with editable NeRFs, \cite{lin2024dynamic} has also discussed a few works in scene editing alongside dynamic NeRFs. Furthermore, \cite{gao2022nerf} discussed many editing applications, including ClipNeRF \cite{CLIP-NeRF}, EditNeRF \cite{Editing_Conditional_RF}, CodeNeRF \cite{codenerf}, CoNeRF \cite{CoNeRF}, however the main focus \cite{gao2022nerf} is oriented towards 3D computer vision applications rather than practical field applications. 

\subsection{Surveys on Gaussian Splatting}

Recent advancements in implicit neural representations have significantly improved the ability to synthesize photorealistic views of complex 3D scenes. Despite their success, NeRF-based methods often suffer from high computational training and rendering costs and require dense, high-quality image datasets for optimal performance. These limitations have spurred interest in alternative representations and optimization techniques. 3DGS has emerged as an alternative explicit scene representation that offers a trade-off between efficiency and fidelity. Unlike NeRFs, which encode a scene in neural network weights, 3DGS represents scenes as a set of 3D Gaussian primitives, each encodes spatial position, size, and opacity, allowing for real-time rendering through forward projection into image space. Since introduced in \cite{3DGS}, 3DGS has demonstrated substantial improvements in rendering speed while maintaining competitive visual quality and relatively straightforward editing capabilities. In an attempt to summarize the huge amount of published works in 3DGS, several recent surveys provide a focused overview of its advancements \cite{luo2024review, chen2024survey, wu2024recent, fei20243d, bao20243d, dalal2024gaussian, bagdasarian20243dgs}. A survey of 3DGS optimization and reconstruction techniques has been the focus of \cite{luo2024review} in which compression \cite{bagdasarian20243dgs}, densification, splitting, anti-aliasing, and reflection enhancement are among the optimization techniques discussed. Furthermore, \cite{luo2024review} summarizes the various methods for mesh surface extraction as well as both object-level and large-scale scene reconstructions. \cite{dalal2024gaussian} has limited the focus of their survey to recent developments in 3D reconstruction techniques in 3DGS. \cite{chen2024survey} is another notable survey on 3DGS that focuses on the practical applicability of 3DGS leveraged by the unprecedented real-time rendering capabilities of 3DGS in applications such as virtual reality and interactive multimedia. Furthermore, \cite{chen2024survey} has also discussed and analyzed the applicability of leading 3DGS models on various benchmarks and evaluated their performance for their utility in these applications. While \cite{chen2024survey} has been dedicated to 3DGS applications, \cite{wu2024recent} has focused on the various 3DGS methods and architectures by classifying existing works into 3D editing and 3D reconstruction. Along with an extensive overview of 3DGS, \cite{bao20243d} has discussed both applications, optimization methods, and extensions of 3DGS. More systematically, the comprehensive survey \cite{fei20243d} provided a unified framework with a novel taxonomy of existing works in 3DGS in six main categories, namely representation, reconstruction, manipulation, generation, perception and virtual humans. Although static and dynamic scene editing has been discussed in the manipulation category of \cite{fei20243d}, its objective is a more comprehensive review rather than a more detailed discussion of a specific topic of 3DGS as the case of the present survey paper that focuses on editing aspect.

Existing surveys on both NeRFs and 3DGS have generally aimed toward a broader and more comprehensive review of either both technologies \cite{tosi2024nerfs} or toward a specific aspect of them. Although an important topic of NeRFs and 3DGS, scene editing has been roughly discussed in most of the general survey papers, and very few works discussed this specific topic in detail \cite{lu2024advances, zhan2023multimodal}. The present survey paper provides a comprehensive survey and taxonomy of the state-of-the-art on both NeRFs and 3DGS with a focus on the different methods and approaches of scene editing as well as their applications. We also shed light on the future challenges and opportunities of potential applications.

\section{Representations of Radiance Field} \label{sec:background}

This section presents an overview of the mathematical foundations of volumetric rendering as well as the implicit and explicit representations of radiance fields using
NeRfs \cite{mildenhall2021nerf} and 3DGS \cite{3DGS}, respectively. These two representations found the basis of the taxonomy of editing approaches of the radiance field introduced in this paper in Section~\ref{sec:taxonomy}. 

\subsection{Volumetric Rendering}

% \textcolor{red}{TODO: Gamal}

While surface rendering considers light interactions with object surface boundaries, volumetric rendering models the light interaction with a medium distributed throughout its volume. This incorporates modeling light absorption, scattering, and emission properties of this medium. In this context, the goal is to calculate a radiance $L(\mathbf{r}, \mathbf{d})$ observed at a point $\mathbf{r} \in \mathtt{R}^3$ along a ray defined by a direction $\mathbf{d} \in \mathtt{R}^3$. The rendering equation used in volumetric rendering can be expressed as an integral of radiance along the ray path:

\begin{equation}
L(\mathbf{r}(t), \mathbf{d}) = \int_{t_n}^{t_f} T(t) \sigma(\mathbf{r}(t)) \mathbf{c}(\mathbf{r}(t), \mathbf{d}) \, dt    
\end{equation}

\noindent where $t_n$ and $t_f$ are the near and far bounds of the ray and $T(t)$ is the transmittance along the ray up to $t$, which represents the fraction of light that survives from point $t_n$ to $t$ without being absorbed or scattered out of the path:
\begin{equation}
T(t) = \exp \left( -\int_{t_n}^{t} \sigma(\mathbf{r}(s)) \, ds \right),    
\end{equation}

\noindent $\sigma(\mathbf{r}(t))$ is the density at point $\mathbf{r}(t)$ and $\mathbf{c}(\mathbf{r}(t), \mathbf{d})$ is the emitted color at point $\mathbf{r}(t)$ in viewing direction \(\mathbf{d}\). Without loss of generality, the problem of scene editing amounts to modifying the radiance field $L(\mathbf{r}, \mathbf{d})$ representing the original scene into a new $L^\prime(\mathbf{r}, \mathbf{d})$ for all $\mathbf{r}$ and $\mathbf{d}$ in this volume.

\subsection{Neural Radiance Fields}

NeRFs \cite{mildenhall2021nerf} use neural networks to build a compact representation of a scene underlying a 3D volume, making it possible to render this volume from novel views. This neural network maps 3D spatial coordinates and viewing directions to color and density values. Formally, NeRF is defined as a function:
\begin{equation}
\mathcal{F}_\theta : (\mathbf{x}, \mathbf{d}) \rightarrow (\mathbf{c}, \sigma)    
\end{equation}
\noindent where \(\mathbf{x} \in \mathbb{R}^3\) is a spatial coordinate, \(\mathbf{d} \in \mathbb{R}^3\) is the viewing direction, \(\mathbf{c} \in \mathbb{R}^3\) is the RGB color, and \(\sigma \in \mathbb{R}\) is the volume density at the point \(\mathbf{x}\). NeRF synthesizes novel views by applying volumetric rendering. The color \(C(\mathbf{r})\) observed along a ray \(\mathbf{r}(t) = \mathbf{o} + t\mathbf{d}\) is computed by integrating the color and density along the ray, weighted by the transmittance \(T(t)\):

\begin{equation}
C(\mathbf{r}) = \int_{t_n}^{t_f} T(t) \sigma(\mathbf{r}(t)) \mathbf{c}(\mathbf{r}(t), \mathbf{d}) \, dt    
\end{equation}

In contrast to traditional volumetric methods that require explicit voxel grids or point clouds, NeRF represents a scene implicitly through the neural network $\mathcal{F}_\theta$ that maps 3D coordinates and viewing directions to color and density values. Minimizing the difference between the rendered color values $C(\mathbf{r}_i)$ and the true pixel colors from the images $C_{\text{gt}}(\mathbf{r}_i)$ typically using a photometric loss:
\begin{equation}
\mathcal{L} = \sum_{i=1}^N \| C(\mathbf{r}_i) - C_{\text{gt}}(\mathbf{r}_i) \|_2^2    
\end{equation}

In this paper, \textit{implicit editing} of radiance fields refers to approaches and methods for modifying the content of the scene by editing $\mathcal{F}\theta$. This is typically achieved by re-training the NeRF model to incorporate changes, resulting in a new model $\mathcal{F}^\prime\theta$.

\subsection{3D Gaussian Splatting}

As outlined in \cite{3DGS}, the 3D volume could be decomposed and represented explicitly using a finite set of \( n \) anisotropic 3D Gaussians, denoted as \(\mathcal{G} = \{ \mathcal{G}_i \}\), for \( i = 1, 2, \dots, n \), each can be mathematically described by a 5-tuple \(\mathcal{G}_i = \langle \mu_i, S_i, R_i, \alpha_i, c_i \rangle\), where, \(\mu_i \in \mathbb{R}^3\) is its mean, \(S_i \in \mathbb{R}^3_{+}\) is the scale vector representing the standard deviation along the principle axes, \(R_i \in SO(3)\) is the rotation matrix, representing the orientation of the corresponding ellipsoid with respect to a world reference coordinate system, \(\alpha_i \in (0,1)\) defines the opacity, and \(c_i \in \mathbb{C}^3\) is a view-dependent color, often expressed as coefficients in a spherical harmonics (SH) basis. The 3D volume occupied by a \(\mathcal{G}_i\) could be described as:

\begin{equation}
G_i(x) = e^{-\frac{1}{2} (x - \mu)^T \Sigma^{-1} (x - \mu)}    
\end{equation}

The covariance matrix \(\Sigma\) of \(\mathcal{G}_i\) is decomposed as:

\begin{equation}
\Sigma = R S S^T R^T    
\end{equation}

For rendering, these 3D Gaussians are projected into 2D, and their covariance matrices are transformed accordingly. This requires computing a new covariance matrix \(\Sigma'\) in the camera's coordinate space, using the Jacobian \(J\) of the affine approximation of the projective transformation, along with a viewing transformation \(W\), as described in \cite{zwicker2001ewa}:

\begin{equation}
\Sigma' = J W \Sigma W^T J^T    
\end{equation}

The color \(c\) of a pixel is calculated by \(\alpha\)-blending the contributions from \(N\) sequentially ordered 2D splats:
\begin{equation}
c = \sum_{i=1}^{N} c_i \alpha_i \prod_{j=1}^{i-1} (1 - \alpha_j)    
\end{equation}

In this paper, \textit{explicit editing} of radiance fields refers to approaches that directly modify the individual Gaussians of $\mathcal{G}$ and their associated parameters to produce an updated model, $\mathcal{G}^\prime$.

\section{Taxonomy}
\label{sec:taxonomy}
\begin{sidewaysfigure}
\centering
\includegraphics[width=\textwidth]{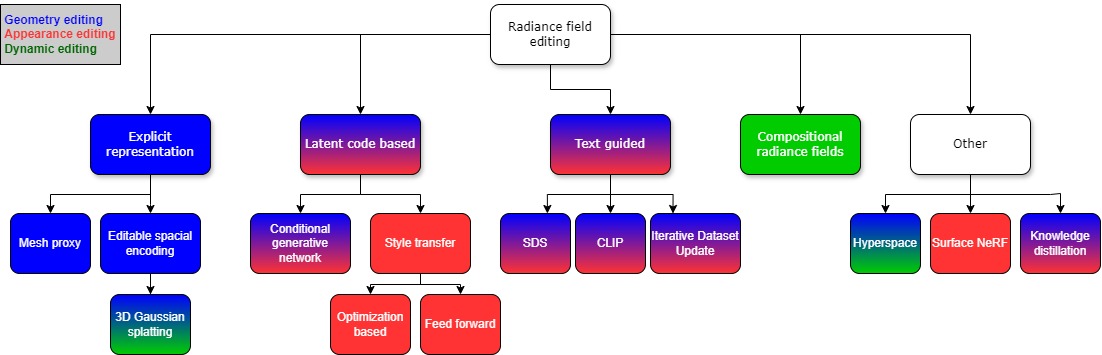}
\caption{Taxonomy of editing methods for radiance fields.}
\label{fig:Taxonomy}
\end{sidewaysfigure}

This paper proposes a novel taxonomy for classifying different types of radiance field editing strategies, organized in a tree-based structure as shown in Fig. \ref{fig:Taxonomy}. Editing strategies and types vary widely, depending on specific goals and applications. We identify three different types, namely \textit{geometry editing}, \textit{appearance editing} and \textit{dynamic editing}. 
Among the categories in our taxonomy, most focus primarily, although not exclusively, on one specific type of editing. In this section, we will explain, for each family, how they achieve the editing of NeRFs, the pros and cons of the desired editing, and details on the most relevant works and models. In Section \ref{sec:application}, we will then showcase popular applications of these methods and compare their performance.

\subsection{Explicit representation}

The challenge of editing NeRFs primarily arises from the implicit nature of their scene representation, which is encoded within the weights of a neural network. This implicit encoding makes it particularly challenging to identify which part of the network to modify in order to edit and deform the geometry and appearance of specific objects in the scene. To address this challenge, a widely adopted solution involves transitioning to an explicit 3D representation that is more amenable to direct editing. This approach can be categorized into two distinct methods.
The first method entails generating an explicit representation, such as meshes derived from a NeRF model, performing the desired edits, and subsequently propagating these changes back into the NeRF model.
The second method involves changing the scene representation to include a more explicit representation, thereby facilitating easier modifications. Notable among these techniques is 3DGS \cite{3DGS}, which we will explore in greater detail due to its growing importance in the context of radiance fields.

\subsubsection{Mesh-based proxy}
\label{mesh}

\begin{figure}[t!]
\centering
\includegraphics[scale = 0.75]{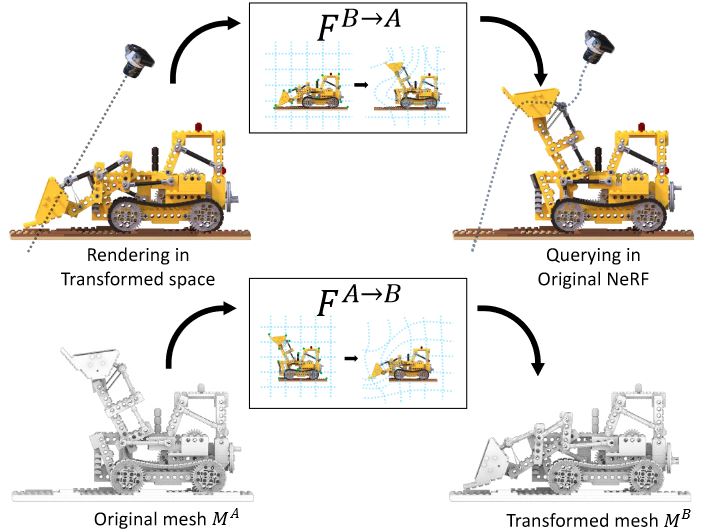}
\caption{Transferring mesh deformation to an implicit radiance field causing ray bending (NeRFDeformer \cite{NeRFDeformer})}
\label{fig:Mesh_def}
\end{figure}

Polygon meshes are an explicit representation widely used due to the availability of numerous 3D modeling tools and techniques that enable straightforward editing, such as Blender \cite{blenders} and ARAP (As-Rigid-As-Possible deformation \cite{ARAP}). In this context, some approaches first construct a NeRF representation and then apply marching cubes \cite{lorensen1998marching} to convert it into a mesh. The key challenge then becomes transferring the edits made on the mesh back to the radiance field. This is typically achieved by computing the displacement between the original and edited meshes and applying this displacement to the sampled ray points in the radiance field. During rendering, the rays are bent accordingly to reflect the desired geometric modifications (see Fig. \ref{fig:Mesh_def}).

In some cases, the reconstruction obtained solely through marching cubes is too coarse to capture fine details. To address this, methods such as those in \cite{interactive}, \cite{Neumesh}, and \cite{NeRF-Editing} built upon the approach introduced by NeuS \cite{wang2021neus}, leverage a neural signed distance function (SDF) to learn object surfaces with a more accurate representation. 

In addition, the methods described in NeRF Editing \cite{NeRF-Editing} convert the user-deformed surface cages into a volumetric tetrahedral mesh. This technique is further refined in \cite{interactive} where a geometry constraint is introduced to the model to better preserve the symmetry of objects. 
Additionally, geometry-aware NeRF editing also allows the model composition of two separate NeRF parts into one. 

Similarly, the methods presented in \cite{Deforming_radiance_cages} and \cite{Nerfshop} adapt cage-based deformation (CBD) \cite{nieto2012cage} to radiance fields by generating coarse 3D mesh cages. 
These cages enable more precise edits of the fine cage surrounding the target object using mean value coordinates (MVC) \cite{ju2023mean}. In the same vein, NeRFshop \cite{Nerfshop} combines tetrahedral meshes with cages to address their individual limitations. 
Specifically, it introduces a pipeline to enhance the mesh generated by marching cubes to avoid relying on the surface representation from NeuS that struggles with intricate details. Additionally, it incorporates a second interpolation step following MVC and employs a NeRF structure similar to instant-NGP \cite{muller2022instant} for faster rendering, enabling real-time interactive editing.
 
\subsubsection{Editable spacial encoding}

Although previous models used an explicit proxy representation to edit an object, the backbone of their model is still the implicit representation of the NeRF. In this section, we will showcase other types of radiance fields that include varying degrees of spatial understanding and explicit representation within the model itself. Many of these models differ from NeRFs since they rely less on a learned implicit representation. Nevertheless, we have chosen to include them in the survey due to their close alignment with the NeRF paradigm, as well as due to their increasing prominence within the field and focus on editing. In particular, while 3DGS \cite{3DGS} qualifies as one such model, it will receive further attention in the next section \ref{3DGS} and throughout this survey due to its high-quality results, its popularity, and its ability to generalize many strategies dedicated to NeRFs. 

\textit{NeuMesh} \cite{Neumesh} decomposes the scene into a mesh, where each vertex learns a local space implicit field, storing geometry and texture information separately. This representation allows for easy geometric editing since the space is locally encoded in the mesh, meaning that any editing to the mesh will directly transfer to the whole 3D space. This disentanglement also allows for separate editing of object appearance using learned latent code (see Section \ref{CGA}). On the other hand, Control-NeRF \cite{Control-NeRF} learns a volumetric representation of multiple scenes at once, while \textit{neural sparse voxel field} (NSVF)\cite{liu2020neural} learns a set of voxel-bounded implicit fields, allowing them to mix different scenes together. Although the NeuMesh and NSVF representations are conceptually similar, NSVF offers much faster inference, while NeuMesh has greater editing power.

\subsubsection{3D Gaussian Splatting} \label{3DGS}

3DGS is gaining popularity within the NeRF community due to its ability to effectively address many of the limitations of NeRF. In particular, the editing limitation of NeRFs is much easier to tackle with this representation because of the explicit representation of the scene. However, it is not trivial either, as it raises its own limitations and challenges, which we will address next.

Likewise NeRF models, different approaches have been proposed for editing scenes represented by 3D Gaussians, sometimes very similar to their NeRF counterparts. As such, many of these methods will be detailed in the corresponding parts of this paper, while this section focuses on the pros and cons of both representations when it comes to scene editing. 

An advantage of 3DGS is demonstrated by GaussianEditor \cite{GaussianEditor_GS}, which introduces hierarchical splatting. This approach tracks the age of Gaussians during the editing process, enabling the adjustment of rigidity or flexibility for specific Gaussian generations. This mechanism facilitates more stable and precise edits.
Moreover, since the Gaussian-based representation is already divided into many small primitives, as will be presented in Section \ref{seg}, it is much easier to separate a scene into different components to manipulate them separately. GaussianGrouping \cite{GaussianGrouping} utilize this advantage to achieve easier object segmentation and move any object around the scene.

\subsection{Latent space based NeRF} \label{latent}
Although explicit representation can allow for very precise editing and provides a good solution, manual editing of complex 3D representations still requires specific 3D graphical skills, making it inconvenient for many users. Another popular family of methods for editing radiance fields is based on latent spaces to include varying degrees of control to NeRFs. We divide this section into two categories. The first category encompasses generative approaches (Section \ref{CGA}) that leverage conditional generative models such as GANs, guided by latent codes. The second is about the application of the style transfer field to radiance fields (Section \ref{ST}).

\subsubsection{Conditional generative approaches} \label{CGA}

\begin{figure}[t]
\centering
\includegraphics[scale = 1.5]{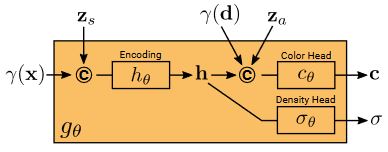}
\caption{Conditional NeRF on shape and appearance code \cite{graf}. The appearance code $z_a$ only affects appearance and not density.}
\label{fig:condNeRF}
\end{figure}

Presented soon after the introduction of the original NeRF paper, Schwarz et al. propose \textit{Generative Radiance Field} (GRAF) \cite{graf}, a model based on Generative Adversarial Network (GAN). GANs are a family of popular architectures for training generative networks that have proven to be very efficient at creating new images. It consists of two networks, a generator and a discriminator, which perform the training task mutually. In GRAF, the generator is based on a conditional radiance field (see Fig. \ref{fig:condNeRF}), taking two latent codes as additional input, one is used for the shape, where the other is used for the appearance. During training, the generator will produce images corresponding to specific views of the generated scene, while the discriminator, a Convolutional Neural Network (CNN), will compare that image against the real corresponding view from the training dataset.

Not only does this model create good 3D object generation results, but also it introduces a model that learns to disentangle shape from appearance, each controlled by distinct latent codes. This method allows for simple appearance editing by changing the appearance code during inference. The conditional NeRF architecture presented in GRAF \cite{graf} introduced two major improvements later used by other methods: disentangling shape and appearance to control one without affecting the other and using latent codes to control NeRF features. The challenge lies in how to guide these latent codes to achieve desired edits. Many following models are inspired by this method and use the latent code conditioning the generative radiance field to provide editing \cite{FED-NeRF, CLIP-NeRF, SINE, codenerf, Editing_Conditional_RF, fenerf, Stylenerf, giraffe, Blending-NeRF}. GIRAFFE \cite{giraffe} directly improves GRAF by two novelties First it adds multiples independent neural representation for multiples objects and backgrounds in the scene (similar to the models described in \ref{seg}), and secondly it replaces the 3D color output of the model with a low resolution and higher-dimensional feature vector. This composite generative feature field not only allows for higher control of different objects and backgrounds within a scene, but also it improves the rendering speed and image quality. EditNeRF \cite{Editing_Conditional_RF} on the other hand adds a shared network to the conditional NeRF, trained on a distribution of specific objects, as well as an instance network specific to the target scene, giving the model strong flexibility to retrain specific parts of the network for editing and maintaining consistency for objects within the distributions. Other disentanglement-based approaches, such as \cite{codenerf} and \cite{Stylenerf}, interpolate latent codes to edit texture or shape. These codes can also be optimized using text-to-image model for editing \cite{CLIP-NeRF,SINE, Blending-NeRF}, which will be seen in greater detail in Section \ref{txt_based}.

\subsubsection{Style transfer} \label{ST}

Style transfer is a popular field of computer vision that aims at transferring artistic features from reference images, generally extracted by a CNN architecture like Visual Geometry Group (VGG) \cite{simonyan2014very}, to different content. 2D style transfer can be separated into two main categories, namely \textit{Optimization-based} and \textit{Feed-forward} style transfer. The former iterates over the target 2D image to optimize a style loss, while the other aims at optimizing a stylization network that captures the style of the reference image and transfers it to the target image in a single pass. Although there are many existing works for 2D style transfer \cite{gatys2016image}, 3D style transfer is a more recent field with few pioneering works like \textit{Learning to Stylize Novel Views} (LSNV) \cite{LSNV}, and even fewer work on NeRFs. Additionally, video style transfer also introduces new challenges, such as consistency over time. Although some methods partially solve the video issue using optical flow or time constraint \cite{wang2020consistent}, they do not apply to 3D style transfer due to inconsistency in novel views. Finally, it is important to note that the following methods focus only on editing appearance style and do not provide geometry or density editing.

\begin{figure}[t]
\centering
\includegraphics[width=\columnwidth]{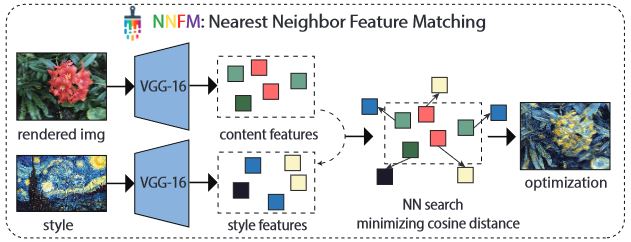}
\caption{Optimization pipeline of ARF \cite{arf} and its NNFM loss.}
\label{fig:arf}
\end{figure}

\begin{itemize}
    \item \textit{Optimization-based:} As one of the first papers on style transfer for NeRF, ARF \cite{arf} introduces Nearest Neighbor Feature Matching (NNFM). This method optimizes the NeRF by extracting feature maps from both the style image and the rendered NeRF (see Fig. \ref{fig:arf}). It then optimizes the smallest cosine distance between both feature vectors at any given point: 
    \begin{equation}
        l_{nnfm}(F_{render},F_{style})= \\ \frac{1}{N}\sum_{i,j}\min_{i',j'}D(F_{render}(i,j),F_{style}(i',j')) 
    \end{equation}

    where $F_{render}$ and $F_{style}$ are the VGG feature maps, respectively extracted from an image rendered by the radiance field, and the style image. Although it produces good stylizing results, this method tends to convert the style of the entire NeRF. Another work CoARF \cite{CoARF} extends this approach by adding semantic understanding using LSeg features \cite{li2022lseg}, allowing for more precise control of which object will be affected. Similarly to approaches such as StyleRF and NeRF-Art \cite{liu2023stylerf, wang2023nerf}, this method also enables compositional style transfer by using different style images and applying their styles to specific parts of the scene individually.

    \item \textit{Feed-forward:} To achieve faster stylization operations, \cite{stylizing_hypernet} trains a hypernetwork to predict the modified weight in the appearance branch of a disentangled NeRF. This network uses a latent vector obtained from the encoder of a variational autoencoder (VAE) that extracts style information from the style image using VGG. Although this method can generalize to any style, it does not manage to replicate complex details from the style of the reference image. On the other hand, \cite{huang2022stylizednerf} achieves a higher quality transfer of style detail by learning the latent codes of given styles. However, they cannot generalize their methods to previously unseen styles. With a different approach, StyleRF \cite{liu2023stylerf} replaces the color output of the NeRF representation with a feature vector. Using Deferred Style Transformation (DST) on the volume-rendered feature of the scene, they add the style information biased by the sum weight of the points along the rays to create the stylized feature map, before using a CNN decoder to generate the final stylized image.
\end{itemize}

Recent works also extend these methods to the 3DGS paradigm \cite{jain2024stylesplat, liu2024stylegaussian, saroha2024gaussian, zhang2024stylizedgs}. \textit{StyleSplat} \cite{jain2024stylesplat} improves \cite{CoARF} using 3D semantic masks on Gaussians. A considerable advantage of optimization-based against feed-forward style transfer is that since they apply to already trained NeRFs, these methods can be implemented on different types of NeRF or 3DGS representation, making them more durable to new advances in the field. However, they also add a potentially long optimization process after the training of NeRFs for every stylization. On the other hand, feed-forward model provides a faster solution. When it comes to the final rendering quality, a quantitative comparison can be difficult. Many models offer hyperparameters to control the degree of stylization, and some recent works also aim to improve local stylization control \cite{zhang2024stylizedgs, li2023arf-plus}. Ultimately, style transfer is a trade-off between original content preservation and new style transfer. It is up to the individual user to decide which model best suits their preferences.

\subsection{Text guided editing} \label{txt_based}

Similarly to our field of interest, the rise of generative image models led to research on diffusion models to generate 3D assets, as well as NeRFs. This section showcases how different types of generative model are used to edit radiance fields with text. We will mention the popular loss introduced by \textit{DreamFusion} \cite{dreamfusion}, as well as the \textit{Contrastive Image Language Pre-training} (CLIP) \cite{radford2021CLIP}. It will also mention the \textit{Iterative Dataset Update}(IDU) strategy as proposed by \cite{haque2023instruct}.

\subsubsection{Score Distillation Sampling} \label{SDS}

\begin{figure*}[t!]
\centering
\includegraphics[width=\textwidth]{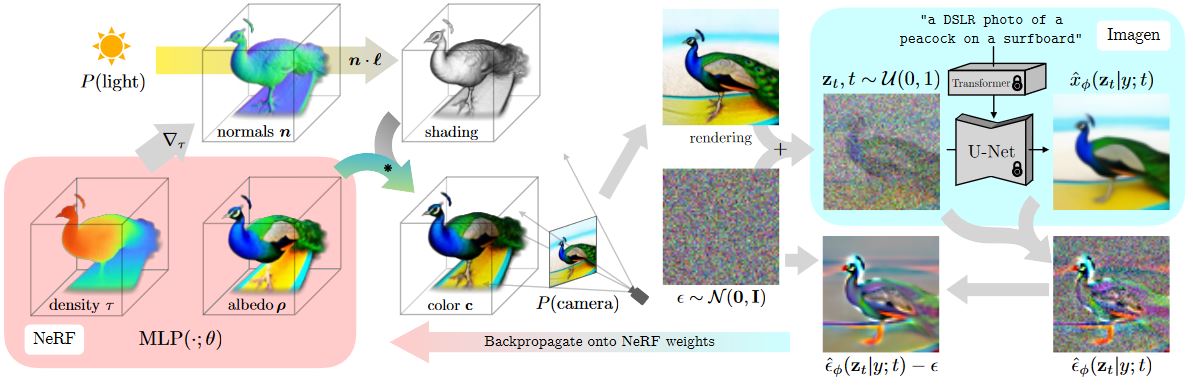}
\caption{3D generative pipeline proposed by DreamFusion \cite{dreamfusion}.DreamFusion diffuses the rendered images and reconstructs it with a conditional Imagen to predict the injected noise \(\hat{\epsilon}_\phi(z_t|y;t)\). The injected noise is then subtracted from and backpropagated to the NeRF.}
\label{fig:DreamFusion}
\end{figure*}

A major work for 3D scene generation with diffusion models is DreamFusion \cite{dreamfusion} which introduced \textit{Score Distillation Sampling} (SDS). This method perturbs rendered images with random noise and feeds them to a diffusion model, which provides a score that guides training (Fig. \ref{fig:DreamFusion}). This SDS loss was consequently used by many different models (cf. \ref{SDS}) to guide NeRF optimization with a diffusion model. It can be noted that this approach is similar to the style transfer approach presented in Section \ref{ST}, although the style reference is not a user-given image but a generative model.

In recent work, many authors have used SDS loss, not only to guide the creation of 3D NeRF assets, but also to edit existing scenes based on a given text query \cite{zhuang2023dreameditor, DynVideo-E}. This method aims at optimizing trained NeRF models using the same SDS-based method to make a given scene closer to the input prompt. DreamWaltz \cite{huang2024dreamwaltz} and DreamHuman \cite{kolotouros2024dreamhuman} explore precise editing of human characters. The approach was further improved by \cite{TELA} that focuses on clothes manipulation and \cite{DynVideo-E} that focuses on human video editing by mixing SDS loss and style transfer methods like NNFM. DreamEditor \cite{zhuang2023dreameditor} encodes the scene in a NeuMesh \cite{Neumesh} representation, then identifies the editing region and optimizes its color, geometry, and vertex position, altering only this region without affecting the background.

\subsubsection{Contrastive Image Language Pre-training}
CLIP \cite{radford2021CLIP} is a very popular method that encodes the image and text pair in a shared latent space. A CLIP loss thus allows for the comparison of two images, two prompts, or an image and a text by calculating their distance within the shared CLIP latent space. It is very useful to guide the latent codes of a conditional NeRF (cf. \ref{CGA}). Works, including \cite{CLIP-NeRF,SINE, Blending-NeRF, Stylenerf}, leverage the CLIP approach to enable text-based editing of a scene. CLIP-NeRF \cite{CLIP-NeRF} in particular uses the CLIP image/text encoder to map the desired text or image input to small steps in the latent space, which are added to the conditional latent codes to deform the NeRF. NeRF-Art \cite{wang2023nerf} on the other hand adapts this method by optimizing a global-local contrastive loss, using a method similar to the NNFM from ARF \cite{arf} (cf. \ref{ST}) to optimize specific features of the original NeRF in the CLIP space with equivalent features of the target prompt.

\subsubsection{Iterative Dataset Update}

Relying on diffusion models as well for editing, some authors \cite{ViCA-NeRF, haque2023instruct} use the \textit{Instruct Pix2Pix} \cite{brooks2023instructpix2pix} model for another strategy. This model allows precise editing of 2D images and it has been extended to 3D radiance fields by propagating edits made to a single or few images to the entire collection of images, which is then used to train the modified NeRF.

Recall that NeRFs are trained using a collection of posed images of a given scene. While many of the previously mentioned methods work on the NeRF architecture, rendering, or scene representation, some authors work on editing the training images directly.

Although not relying on diffusion models at the time, SNeRF \cite{nguyen2022snerf} initially proposed a style transfer application to train a NeRF model and update training images simultaneously on which the model is trained. While guiding the NeRF training with edited images can intuitively allow change in the model, guiding the image editing with the NeRF training provides view consistency when propagating edits. Following this example, Instruct NeRF2NeRF \cite{haque2023instruct} leverages instruct Pix2Pix \cite{brooks2023instructpix2pix} to gradually edit viewpoints in the collection by replacing training images with rendered images edited by Pix2Pix, leading to any view inconsistency in the dataset to disappear over training time. ViCA-NeRF \cite{ViCA-NeRF} in particular focuses on improving the view consistency from \cite{haque2023instruct} by adding two sources of regularization: one leveraging the geometric information form NeRF and the other learned by aligning the Instruct Pix2Pix latent code from edited and unedited images.

\subsection{Compositional Radiance Fields} \label{seg}
\begin{figure}
    \centering
    \includegraphics[width=\linewidth]{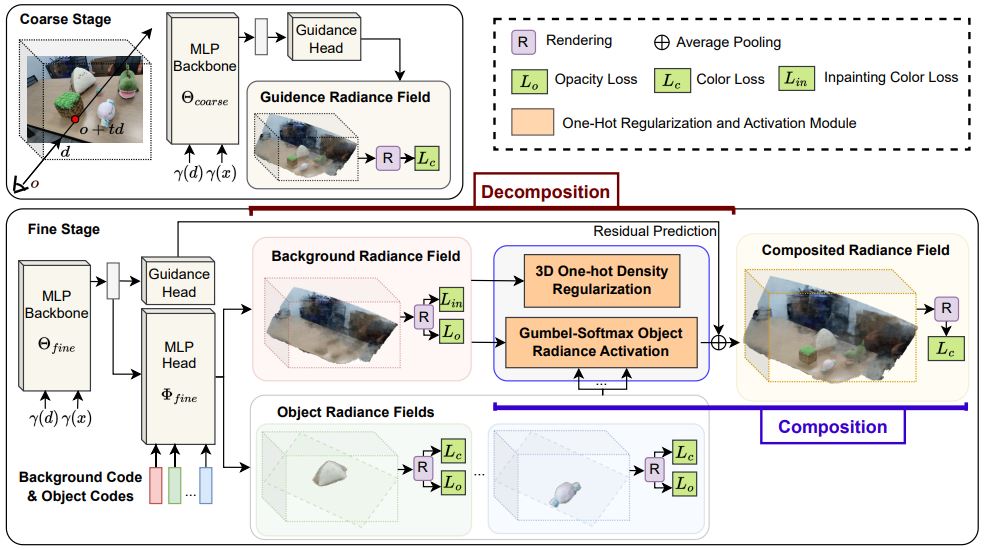}
    \caption{Unified compositional editing of NeRFs \cite{wang2023learning}.}
    \label{fig:compositional-nerf}
\end{figure}

To avoid the challenges of directly editing radiance fields, several approaches focus on decomposing the scene into multiple models, each corresponding to an object or feature in the scene. Applying modifications to these individual models and then using compositional rendering (see Fig. \ref{fig:compositional-nerf}) allows to generate newly edited scenes.  
These methods aim to create distinct, independent models for individual objects within a scene, enabling their relative movement or transformation. The primary challenge in these approaches is effectively decoupling dynamic objects from the static background.
Although NeRFs and 3DGS show promise for static scenes, they face significant challenges in modeling and editing dynamic environments with moving objects, changing lighting conditions, and exposure variations \cite{dahmani2024swag, martin2021nerf}. Early efforts to address these limitations have focused on removing transient objects from the scene and reconstructing only the static components. However, this approach often introduces artifacts resulting from the presence of transient objects \cite{guo2023streetsurf, dahmani2024swag, martin2021nerf, tancik2022block}.

Alternative approaches aim to represent dynamic scenes either by employing time-varying radiance fields \cite{turki2023suds, yang2023emernerf, nguyen2024rodus} or by using scene graphs to define the relationships between static and dynamic objects over time \cite{mars, yan2024street, DrivingGaussian, ost2021neural, mohamad2024denser, chen2024omnire, peng2024desire, zhou2024hugs}. However, scene graph-based models frequently struggle to accurately capture the time-varying appearances and deformations of dynamic objects.

Dynamic scene representations have also evolved with the introduction of 4D neural scene representations, which treat time as an additional dimension alongside spatial dimensions to capture single-object dynamics \cite{attal2023hyperreel, fridovich2023k,park2021hypernerf}.

\subsection{Other} \label{other}
\subsubsection{Hyper-space}\label{hyper}

To deal with the issue of NeRF editing, many models are inspired by another research domains in NeRF: dynamic NeRF. As many dynamic NeRF models rely on a deformation field to represent the temporal changes in a scene, some models use the same strategy to represent user changes.

Following the strategy of HyperNeRF \cite{park2021hypernerf}, it is possible to represent the input images as slices of a \textit{hyper-space}. The main advantage of this space is that, unlike other deformation fields, it is capable of handling discontinuous deformation, such as mouth opening, objects attaching and detaching, with a continuous function such as a Multi-Layer Perception (MLP) by changing coordinates inside the hyper-space \cite{park2021hypernerf}. Although HyperNeRF does not support editing, EditableNeRF \cite{Editablenerf} and CoNeRF \cite{CoNeRF} both build on its method, allowing them to edit dynamic scenes, but also to handle discontinuous deformation. These models introduce key points in the scenes that can be manually entered \cite{CoNeRF} or automatically detected \cite{Editablenerf}. CoNeRF requires users to draw masks on input images, while EditableNeRF detects relevant key points in the scene automatically. It uses depth maps and optical flow acquired through HyperNeRF \cite{park2021hypernerf} and RAFT \cite{teed2020raft} in the input images. They will then train an MLP that associates every point in the scene with the key point that will affect it most. Changes to the key points will be transferred to the video rendering by moving the scene inside the hyperspace coordinates. The EditableNeRF approach \cite{Editablenerf} also improves on \cite{CoNeRF} by improving the user interface. Whereas the latter requires manual labeling and modifying key points by changing scalar values, EditableNeRF allows for direct drag-and-drop manipulation of predefined key points.

%TeTWild
\subsubsection{Surface NeRF}
Popular NeRF representations focus on representing object surfaces of scenes represented by NeRFs. This allows for, not only higher quality rendering and reduces local artifacts, but also allows some models to offer small appearance edits \cite{Nerfactor}. This is especially relevant to simulate different time or weather conditions for a given scene or to edit lighting conditions \cite{martin2021nerf, wang2021neus, srinivasan2021nerv, boss2021nerd}. 

\subsubsection{Knowledge distillation}
Finally, knowledge distillation can be used for NeRF editing with a similar idea as the mesh proxy described earlier, but using another model as a proxy instead of an explicit representation. Seal-3D \cite{Seal-3d} and Seal-ID \cite{SealD-NeRF} train an initial teacher NeRF model of the scene and implement a few editing operations that map an edited space to the source space, then train a student model supervised by a teacher model. The student training adopts a two-stage training, with a pre-training only on local space, allowing the user to get a pre-visualization within seconds.

\section{Applications} \label{sec:application}
In this section, we will focus on the most popular applications for NeRF editing. These include simulation of outdoor synthetic environments, automobile, human motion, body, and face control, as well as artistic stylization and asset generation. For the latter, we will focus mainly on asset generation, as the stylizing process of existing scenes has been extensively covered in Sections \ref{ST} and \ref{txt_based}. We also provide comparison between the most popular models for different tasks in Table \ref{table:comparison}. This comparison is centered on the type of editing method used and the degree of editing control provided.

\subsection{Artistic NeRF}

With the many improvements to NeRF, the state-of-the-art of radiance fields has reached a point where it can be used by the general public, with mobile applications such as \cite{Scaniverse} allowing anyone to create and edit radiance field models easily. As NeRF applications for mobile and virtual reality (VR)  become more widespread, editing radiance fields becomes more interesting for creative expression. The wide variety of filters and special effects in graphics applications that can be added to real scenes using text prompts \cite{CLIP-NeRF, wang2023nerf, GaussianEditor_txt} or reference style \cite{Stylenerf, CoARF, arf} provide great opportunities for augmented reality.
Several works as DreamFusion \cite{dreamfusion} have focused on creating NeRF models directly from a text prompt to generate 3D objects \cite{Interactive3D, huang2024dreamwaltz, kolotouros2024dreamhuman, jun2023shap}. These models can generally be edited again by adding new prompts and continuing the optimization process, providing great flexibility to create controllable assets \cite{Interactive3D} which could later be used directly in VR or video games. Also, works such as SINE \cite{SINE} have successfully combined mesh-based editing and CLIP models to provide very high control of real objects represented by radiance fields.

\begin{sidewaystable}[]
\caption{Comparison of a few popular radiance field editing models}
\label{table:comparison}
\begin{tabular}{|p{2.5cm}|p{2.4cm}|p{9cm}|p{2.5cm}|}
\hline
\hline
\textbf{Models}       & \textbf{Editing Type}                                     & \textbf{Editing Control}                                                                     & \textbf{Video Editing}\\ 
\hline
\hline
StyleRF \cite{liu2023stylerf}           & Style transfer                                            & Editing affects the whole scene, allows style composition                     &   \ding{55}                     \\ \hline
StyleSplat \cite{jain2024stylesplat}            & Style transfer                                            & Can edit specific objects                                                                        & \ding{55}                     \\ \hline
DynVideo-E \cite{DynVideo-E}           & Style transfer                                            & Editing affects the whole scene                                                                                  & \checkmark                    \\ \hline
NeRF-ART \cite{wang2023nerf}             & CLIP                                                      & Editing affects the whole scene                                                                                  & \ding{55}                     \\ \hline
SINE \cite{SINE}                 & CLIP \& Mesh      & Can edit the appearance and geometry of specific objects \& object insertion & \ding{55}                     \\ \hline
NeRFDeformer \cite{NeRFDeformer}         & Mesh                                                      & Can edit the geometry of specific objects                                                                                  & \ding{55}                     \\ \hline
Gaussian Editor \cite{GaussianEditor_txt}& Text                                                      & Can edit the appearance of specific objects                                                                      & \ding{55}                     \\ \hline
Gaussian Editor \cite{GaussianEditor_GS} & Text                                                      & Can edit the appearance of specific objects \& object removal and insertion       & \ding{55}                     \\ \hline
EditableNeRF \cite{Editablenerf}         & Hyperspace                                                & Can edit object geometry with topology changes                                                                     & \checkmark                    \\ \hline
RigNeRF \cite{athar2022rignerf}              & Hyperspace  \& 3DMM & Can edit the expression of human faces with topology changes        & \checkmark                    \\ \hline
StreetGaussian \cite{yan2024street}       & Composition                                               & Object translation \& rotations                                                                         & \checkmark   \\      \hline
\hline
\end{tabular}
% %

% \vspace{0.5em}
\end{sidewaystable}

\subsection{Outdoor synthetic environment simulation}
Large outdoor environment editing is relevant for digital twins \cite{liu2024citygaussian, tancik2022block} and autonomous driving applications. \cite{mars,DrivingGaussian,mohamad2024denser} create a NeRF or 3DGS representation from the recorded driving scenarios and edit the actors in the scene, allowing the creation of new artificial scenarios. This is particularly relevant for an autonomous driving scenario generation where collecting high-quality real data for specific safety-critical scenarios can be challenging. Focusing on object surfaces and lighting conditions, \cite{li2023climatenerf, boss2021nerd, srinivasan2021nerv} have shown the possibility of editing weather conditions in a scene, while NeRF in the wild \cite{martin2021nerf} and urban radiance field \cite{Urban_Radiance_Field_with_DNMP} remove transient occluding objects. 

\subsection{Human editing}

Radiance field editing has become particularly popular for applications involving human body and facial models, leveraging various editing techniques.
Appearance editing includes mixing facial features \cite{Stylenerf, FED-NeRF}, incorporating new features from text embeddings \cite{CLIP-NeRF}, or using diffusion models \cite{ViCA-NeRF}.
Geometry editing can be used to modify facial expressions \cite{athar2022rignerf}, body pose \cite{peng2021animatable} or human movement in a video \cite{Editablenerf}, both of which can be extended with human body parametric models \cite{loper2023smpl}. 
Other works identify latent facial features to achieve realistic head motion \cite{FED-NeRF, fenerf,Stylenerf}. FENeRF \cite{fenerf} adds a semantic mask output to the radiance and density output of the NeRF, then renders both real images and semantic images, training a discriminator on each to enable semantic-guided editing. StyleNeRF \cite{Stylenerf} uses a style-based generator inspired by StyleGAN2 \cite{karras2020analyzing} that maps input latent codes to the style space, allowing for high-quality image rendering, style mixing, and interpolation of faces. RigNeRF \cite{athar2022rignerf} leverage 3D Morphable Face Models (3DMMs) \cite{blanz2023morphable} to guide the facial expression of a video, leading to high-quality head control.

\section{Evaluation} \label{sec:evaluation}
Most of the metrics and datasets used to evaluate NeRF editing are in line with the commonly used evaluation methods for standard NeRF, dynamic NeRF, and 2D graphics. Although they provide useful information on the final rendering quality, they fail to quantify important 3D requirements like view consistency or preservation of unedited background. At the time of writing this survey, there is no consensus on quantitative evaluation metrics that assess the editing capabilities of a model. In this section, we will present evaluation metrics proposed by many authors in this field, along with a review of the classic datasets commonly used for NeRF.

\subsection{Datasets}

\begin{sidewaystable}[]
\caption{List and comparison of popular datasets for radiance field evaluation. This comparison is based on whether the scenes therein are real synthetic or dynamic and how many scenes exist in each category. Further, the number of views per scene, the resolution of each view, and whether a point cloud exists are also considered. Finally, which specific application the dataset is used in is mentioned when it is relevant.}
% \resizebox{\textwidth}{!}{
\begin{tabular}{|l|l|l|l|l|l|l|l|l|}
\hline \hline
\textbf{Dataset}                                                                        & \begin{tabular}[c]{@{}l@{}}\textbf{Real }\\ \textbf{Scenes} \end{tabular} & \begin{tabular}[c]{@{}l@{}}\textbf{Synthetic} \\ \textbf{Scenes}\end{tabular} & \begin{tabular}[c]{@{}l@{}}\textbf{Dynamic} \\ \textbf{Scenes} \end{tabular} & \begin{tabular}[c]{@{}l@{}}\textbf{Views/Scene} \end{tabular}                                                                      & \textbf{FPS}                                                                        & \textbf{Resolution}                                                     & \begin{tabular}[c]{@{}l@{}}\textbf{Point} \\ \textbf{Cloud}\end{tabular} & \textbf{\begin{tabular}[c]{@{}l@{}}Specific \\ Application \end{tabular}} \\ \hline
\begin{tabular}[c]{@{}l@{}}NeRF \\\cite{mildenhall2021nerf}\end{tabular}             & 8                                                     & 12                                                         & \ding{55}                                                        & \begin{tabular}[c]{@{}l@{}}20 - 62\\ (real scenes)\end{tabular}                                                          & n/a                                                              & 1008x756                                                               & \ding{55}              & -                     \\ \hline
\begin{tabular}[c]{@{}l@{}}KITTI \\\cite{KITTY}\end{tabular}           & 22                                                   & 0                                                          & \checkmark                                                      & 4                                                                                    & 10                                                                                                                          & 1383x512                                                          & \checkmark            & Autonomous driving    \\ \hline
\begin{tabular}[c]{@{}l@{}}D-NeRF \\\cite{pumarola2021d}\end{tabular}            & 0                                                     & 8                                                          & \checkmark                                                      & 100-200 views                                                                        & n/a                                                                                                                                                 & 800x800                                                                & \ding{55}              & Dynamic NeRF           \\ \hline
\begin{tabular}[c]{@{}l@{}}Instruct NeRF2NeRF \\\cite{haque2023instruct}\end{tabular}   & 6                                                     & 0                                                          & \ding{55}                                                        & 50-300                                                                               & n/a                                                                                   & 1008x756                                                                      & \ding{55}              & -                     \\ \hline
\begin{tabular}[c]{@{}l@{}}mip-NeRF 360 \\ \cite{barron2021mip}\end{tabular}        & 9                                                     & 0                                                          & \ding{55}                                                        & 100-330                                                                       & n/a                                                                                                                                             & 1.0-1.6 Mp                                                      & \ding{55}              & -                     \\ \hline
\begin{tabular}[c]{@{}l@{}}Plenoptic Video Dataset\\\cite{Neural3D}\end{tabular}  & 6                                                     & 0                                                          & \checkmark                                                      & 1                                                                                    & 30                                                                                                                                     & 2028x2704                                                                      & \ding{55}              & Human movements       \\ \hline
\begin{tabular}[c]{@{}l@{}}Panoptic Dataset\\ \cite{joo2015panoptic}\end{tabular}      & 65                                                    & 0                                                          & \checkmark                                                      & \begin{tabular}[c]{@{}l@{}}480 VGA camera, \\ 30+ HD camera \end{tabular} & \begin{tabular}[c]{@{}l@{}}25\\ 30 \end{tabular}                                                                                                                                              & \begin{tabular}[c]{@{}l@{}}640x480 (VGA)\\ 1920x1080 (HD)\end{tabular} & \ding{55}              & Human movements       \\ \hline
\begin{tabular}[c]{@{}l@{}}Tanks and temples\\\cite{Knapitsch2017}\end{tabular} & 14                                                    & 0                                                          & \ding{55}                                                        & 4,395 - 21,871                                                                      & n/a                                                                                                                                             & 1920x1080                                                              & \checkmark            & Outdoor environments                     \\ \hline
\begin{tabular}[c]{@{}l@{}}Waymo Open \\ \cite{Sun_2020_CVPR}\end{tabular}      & 1150                                                  & 0                                                          & \checkmark                                                      & 5       &10                                                                         & 1920x1280                               & \checkmark              & Autonomous driving       \\ \hline \hline
\end{tabular}

\label{table:dataset}
\end{sidewaystable}

Numerous popular datasets have been introduced specifically for NeRF evaluation and computer graphics, which have gained significant traction in this field. The most popular remains the synthetic dataset from Mildenhall et al. \cite{mildenhall2021nerf} in the original NeRF paper \cite{mildenhall2021nerf}, which is often used to benchmark novel proposals. However, it is quite limited due to its size, lack of dynamic scenes, human motion, or more complex objects, leading to publication of other datasets for different use cases.
Table \ref{table:dataset}  provides a description and comparison of multiple relevant datasets used in the literature, as well as some of the interesting specificity they provide. For static scenes, the views per scene refer to the number of images available for each scenes in the dataset. For real dynamic scenes, it refers to the number of cameras used to capture the scene at the indicated frequency (FPS) and thus the number of view at any given point in time. D-NeRF \cite{pumarola2021d} is the only exception where the synthetic dynamic dataset contains one camera moving across either 100 or 200 views depending on the scene.

\subsection{Metrics}
A common evaluation method for NeRF editing consists in first making the edit and then using classic NeRF evaluation metrics on the edited NeRF to evaluate the final rendering quality. Most common are \textit{Peak Signal to Noise Ration} (PSNR), \textit{Structural Similarity Index Measure} (SSIM)) \cite{wang2004image} and \textit{Learned Perceptual Image Patch Similarity} (LPIPS) \cite{zhang2018unreasonable}, which evaluate the visual quality of the image. For text-based editing in particular, it is common to rely on user studies, where participants rate editing results based on various text prompts to decide which image fits the given prompt and looks best.

To quantify editing quality, CLIP-NeRF \cite{CLIP-NeRF} uses the \textit{Fréchet Inception Distance} (FID) \cite{heusel2017gans} to evaluate the image quality of reconstructed views both before and after editing. Text-based models often calculate the \textit{CLIP score} between the edited image and the text prompt which estimates their similarity in the latent CLIP space \cite{radford2021CLIP}. Some use \textit{Manipulative Precision} (MP) \cite{li2020manigan} that includes the CLIP directional score and the \textit{L1 normalized pixel distance} between the original and edited image for content preservation(see \cite{Blending-NeRF}). The Dynamic 3D Gaussian \cite{luiten2024dynamic} uses 2D long-term point tracking metrics presented in \cite{zheng2023pointodyssey} and adapts them to the 3D space.

Finally, editing time remains an important comparison metric, especially for all optimization-based editing strategies that require a long optimization process.

\subsection{Benchmarks}
Currently, there is a lack of commonly used benchmarks for NeRF editing. Although some metrics and datasets have become very popular for NeRF evaluation, none of them provide specific evaluation criteria for editing. So far, NeRF editing still mostly relies on subjective visual evaluation, focusing on the model's editing capabilities while ensuring it maintains strong performance according to established NeRF metrics. Recently, \cite{NeRFDeformer} proposed a specific dataset to evaluate NeRF editing, but so far it has not gained widespread adoption.

\section{Future research direction}\label{sec:future_works}
\subsection{Challenges}
\textbf{Lack of consensus for reference datasets and metrics:} 

The evaluation of radiance field editing currently lacks standardized datasets and robust metrics tailored to 3D-specific challenges. Most widely used metrics are derived from 2D graphics, making it challenging to assess critical 3D aspects such as view consistency. This limitation is particularly pronounced in appearance editing, where quality assessments often rely on user studies, which are subjective and prone to bias. Furthermore, existing metrics fail to accurately capture the extent to which edits are localized to specific regions of interest without inadvertently altering other parts of the scene. Developing comprehensive benchmarks and metrics that address these shortcomings is essential for advancing the field.

\textbf{User-friendly interfaces:} 

Radiance field editing models offer a range of interfaces for controlling the editing process, including text prompts, reference images, control points, and integration with mesh-compatible 3D editing tools like Blender. Despite these advancements, there is significant room for improvement in terms of precision and controllability. For instance, ensuring that edits are confined strictly to the target regions of interest in the model without affecting other parts of the scene remains a persistent challenge. Refining these interfaces to provide intuitive, accurate, and localized control will be key to enhancing their usability and effectiveness.

\textbf{Enhancing geometry editing beyond specific categories:}
There are many different strategies for geometric editing, but the highest quality models are assisted by specific physical models such as human expressions \cite{athar2022rignerf} or trained on specific categories of objects (such as editing only chairs \cite{Editing_Conditional_RF}). Thus, the variety of objects and scenes that can be edited using these methods is limited to the amount of data or physical models available to guide them. 

However, these methods are inherently limited in scope, as their general performance depends merely on the availability of relevant datasets or physical models to guide the editing process. This restricts their applicability to a predefined set of object types. Extending these techniques to edit previously unseen or unmodeled objects remains a significant challenge, since it requires a more generalized and unified frameworks that are capable of handling diverse and unstructured geometric representations without domain-specific guidance.

\textbf{Reducing editing time:} 

Although ray-bending techniques provide faster alternatives, most editing models continue to rely on lengthy optimization processes for performing even a single edit. This limitation poses a significant challenge, particularly if optimization-based methods remain a popular approach. Addressing this bottleneck will be critical to improving the practicality and scalability of radiance field editing, especially for real-time or interactive vision and graphics applications.

\subsection{Opportunities}
\textbf{Many potential applications:} Although many applications have been explored for radiance field representations, there has been less interest in applications to specifically edit them. Data augmentation, synthetic dataset generation, and mapping offer great opportunities. Artistic applications for asset creation could be very valuable for augmented reality, video games, or special effects in movies. Precise geometry editing tools could be very useful for video editing or photorealistic simulation.  

\section{Conclusion}\label{sec:conclusion}

In this survey, we explored a diverse array of strategies developed to address the challenge of editing radiance fields, proposing a novel taxonomy to classify these approaches. Techniques for editing radiance fields span a broad spectrum, from directly manipulating latent codes and optimizing model parameters to deforming the rendering process. Others have significantly altered the foundational NeRF representation, giving rise to alternative paradigms such as 3DGS and other radiance field frameworks.
The integration of generative AI technologies, particularly text-to-image models, has further expanded the accessibility and applicability of radiance field editing, enabling innovative solutions and attracting a wider user base. These advancements highlight the growing potential of 3D environments as the next frontier in digital content creation, following the rapid progress in image and video generation.
Despite the substantial progress since the introduction of NeRF in 2020, many challenges remain. These include improving editing precision, reducing computational overhead, and expanding the range of editable scenes and objects. Future research should also focus on establishing standardized benchmarks, enhancing user-friendly interfaces, and exploring novel applications across domains such as gaming, virtual reality, and autonomous systems. The continued development of radiance field editing promises to unlock transformative possibilities for 3D content creation and manipulation.

% \nocite{*}
% \bibliographystyle{plain}
% \bibliographystyle{chicago}
\bibliography{sn-bibliography}

\end{document}